\PassOptionsToPackage{dvipsnames}{xcolor}
\documentclass{article} %
\usepackage{iclr2024_conference,times}

\usepackage{amsmath,amsfonts,bm}

\def\eqref#1{equation~\ref{#1}}

\def\1{\bm{1}}

\DeclareMathAlphabet{\mathsfit}{\encodingdefault}{\sfdefault}{m}{sl}
\SetMathAlphabet{\mathsfit}{bold}{\encodingdefault}{\sfdefault}{bx}{n}

\usepackage{hyperref}
\usepackage{url}
\usepackage{microtype}
\usepackage{graphicx}
\usepackage{subfigure}
\usepackage{booktabs} %
\usepackage{xcolor} 
\usepackage[dvipsnames]{xcolor} 

\usepackage[textsize=tiny]{todonotes}

\usepackage[capitalize,noabbrev]{cleveref}

\usepackage{amsmath}
\usepackage{amssymb}
\usepackage{mathtools}
\usepackage{amsthm}

\usepackage{microtype}
\usepackage{graphicx}
\usepackage{subfigure}
\usepackage{booktabs} %
\usepackage[dvipsnames]{xcolor} 
\usepackage{inconsolata}
\usepackage{tabularx}
\usepackage{multirow}
\usepackage{arydshln}
\usepackage{latexsym}
\usepackage{enumitem}
\usepackage{algorithm} 
\usepackage{titlesec}
\usepackage{wrapfig}
\usepackage{algpseudocode}

\titlespacing{\paragraph}{%
  0pt}{%
  0.01 \baselineskip}{%
  1em}%

\newcommand{\linecomment}[1]{\textcolor{blue}{ \(\triangleright\) #1}}
\newcommand{\inlinecomment}[1]{\textcolor{blue}{ \hspace{1em} \(\triangleright\) #1}}

\newcommand{\up}[1]{\textcolor{OliveGreen}{\small \ $\uparrow${#1}}}
\newcommand{\downbad}[1]{\textcolor{Maroon}{\small \ $\downarrow${#1}}}
\newcommand{\down}[1]{\textcolor{OliveGreen}{\small \ $\downarrow${#1}}}

\newcommand{\decode}{\texttt{Activation Decoding}}
\newcommand{\sharpness}{in-context sharpness}
\newcommand{\metric}{entropy}
\newcommand{\representation}{inner representations }
\newcommand{\counterfact}{\textsc{CounterFact}}

\newcommand{\method}[1]{Activation Decoding }

\theoremstyle{plain}

\theoremstyle{definition}

\theoremstyle{remark}

\title{In-Context Sharpness as Alerts: An Inner Representation Perspective for Hallucination Mitigation}

\author{Shiqi Chen\thanks{Equal contribution. Work done during SC's visit to HKUST.} \\
City University of Hong Kong\\
\texttt{schen438-c@my.cityu.edu.hk} \\
\And
Miao Xiong\footnotemark[1] \\
National University of Singapore \\
\texttt{miao.xiong@u.nus.edu} \\
\And
Junteng Liu \\
Shanghai Jiao Tong University \\
\texttt{vep123@sjtu.edu.cn} \\
\And
Zhengxuan Wu\\
Stanford University \\
\texttt{wuzhengx@stanford.edu} \\
\And
Teng Xiao\\
Penn State University \\
\texttt{tengxiao@psu.edu} \\
\And
Siyang Gao\\
City University of Hong Kong \\
\texttt{siyangao@cityu.edu.hk} \\
\And
Junxian He\\
Hong Kong University of Science and Technology \\
\texttt{junxianh@cse.ust.hk} \\
}

\iclrfinalcopy %
\begin{document}

\maketitle
\begin{abstract}
\vspace{-2mm}
Large language models (LLMs) frequently hallucinate and produce factual errors, yet our understanding of why they make these errors remains limited. In this study, we delve into the underlying mechanisms of LLM hallucinations from the perspective of \emph{inner representations}, 
and discover a salient pattern associated with hallucinations: correct generations tend to have \emph{sharper} context activations in the hidden states of the in-context tokens, compared to the incorrect ones. 
Leveraging this insight, we propose an entropy-based metric to quantify the ``sharpness'' among the in-context hidden states and incorporate it into the decoding process to formulate a constrained decoding approach.
Experiments on various knowledge-seeking and hallucination benchmarks 
demonstrate our approach's consistent effectiveness, for example, achieving up to an 8.6 point improvement on TruthfulQA. 
We believe this study can improve our understanding of hallucinations and serve as a practical solution for hallucination mitigation. Code is publicly available at \href{https://github.com/hkust-nlp/Activation\_Decoding}{https://github.com/hkust-nlp/Activation\_Decoding}.
\end{abstract}

\section{Introduction}
\label{sec:intro}
\vspace{-2mm}
Large language models (LLMs) have made remarkable advancements in recent years,
with extensive applications across various domains~\citep{openai2022intro,openai2023gpt,llmchallenges}. Despite these advances, LLMs still face notable challenges regarding factuality, which could critically undermine the trustworthiness and reliability of LLMs, as highlighted in recent studies~\citep{chen2023felm,ji2023survey,wang2023survey}. 
To address the factuality issue, many efforts have focused on retrieving external knowledge~\citep{incontext_rag,yu2023improving, activerag} for generation or fact-checking, as well as fine-tuning~\citep{selfrag} and self-evaluation~\citep{selfcorrection,uncertainty}. However, these methods often require high computational resources or high-quality knowledge bases, which may not be available for domain-specific cases. In contrast, we aim to tackle this challenge from the perspective of model's inner representations, investigating whether the hidden states contain information about hallucination.

To gain this mechanistic understanding of hallucinations\footnote{Hallucinations can be described as outputs that do not conform to the model's inner belief~\citep{zou2023representation}. However, the concept of ``inner belief'' and its measurement remains debatable. Thus in this paper we use ``hallucination'' to denote generation that is not aligned with world knowledge to simplify the discussion, which shares the same definition as ``factuality''.} through the lens of hidden states, 
we begin by formulating an internal knowledge extraction process following~\citet{geva2023dissecting}, to examine whether the inner hidden states of the model contain the target knowledge (which is sometimes described as the model's inner belief). The underlying hypothesis is that hallucinated knowledge may not be encoded into the intermediate hidden states.
Specifically, for example, in the prompt `\texttt{Beats Music is owned by}', if the token `\texttt{Apple}' is encoded within the representations of the subject `\texttt{Beats Music}', we posit that the model ``knows'' \texttt{Beats Music} and \texttt{Apple} are related in some ways and \texttt{Apple} is more likely to be a correct answer. 
In this case, we consider the token `\texttt{Apple}' activated by `\texttt{Beats Music}'. 
Our case study on the \counterfact~dataset~\citep{kevin2022locate} (\cref{tab:sub-ans}) -- where the subject, object, and relation annotations are available -- reveal that the correct answers have significantly higher rate of activation. 

\begin{figure*}[t]
  \centering\includegraphics[width=1\textwidth]{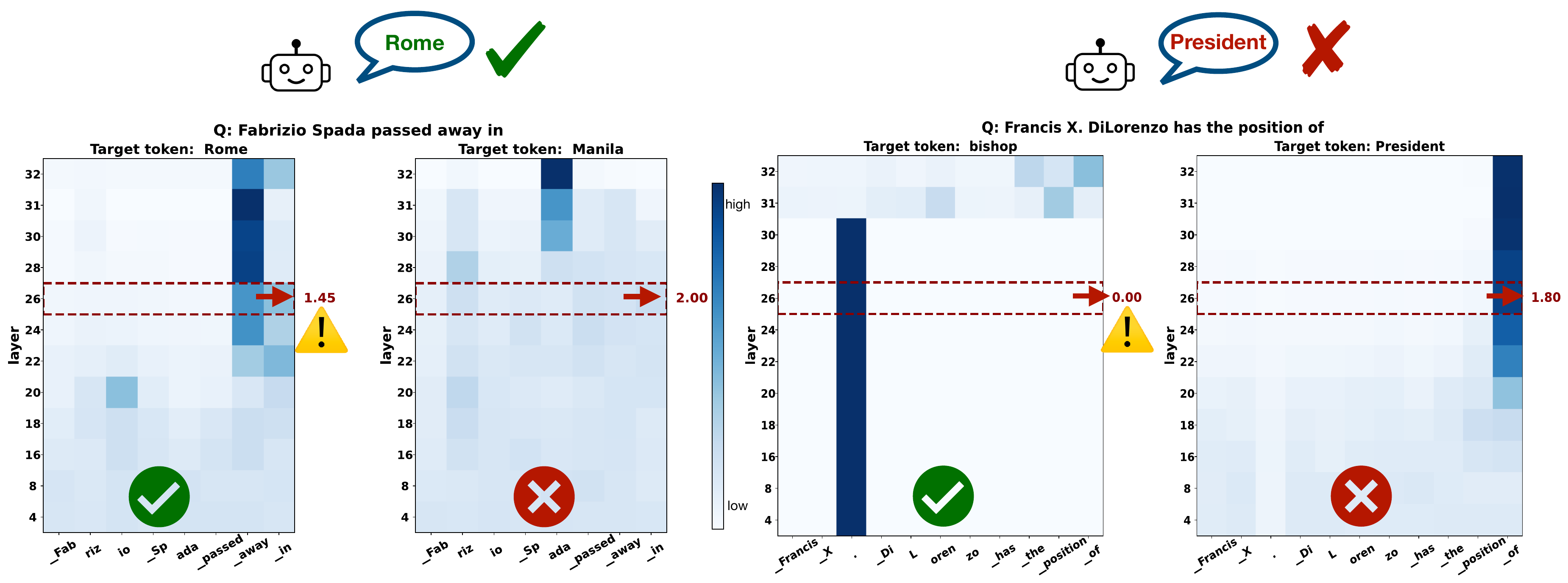}
  \caption{Visualization of why in-context activation can serve as an alarming signal for factuality. For a given question (e.g., ``Fabrizio Spada passed away in \_\_"), we visualize the activation of the truth and false tokens across transformer layers. \textbf{Left}: we use the ground truth and false answers from \counterfact{} (e.g., ``Rome" or ``Manila") as the target tokens. In this example, the model generates the correct answer. \textbf{Right}: we use the ground truth answer and the model's generated false answer. 
  We then calculate the activation entropy across intermediate layers, focusing on the $26$-th layer's entropy value (detailed calculation in \cref{sec:finding2}). 
  This entropy metric is annotated in the figure.
  Our findings reveal that incorrect tokens generally exhibit higher entropy than correct ones.
  }
  \vspace{-4mm}
  \label{fig:box}
\end{figure*}

However, this preliminary study based on the method in~\citet{geva2023dissecting} requires knowledge triplet annotations. To mitigate this requirement, 
we extend our investigation to examining the activations of correct versus incorrect answers across the full input sequence, moving beyond merely the subject focus. Our findings reveal that \emph{correct generations typically exhibit sharper context activations compared to incorrect ones, particularly within the intermediate layers across the in-context tokens.}
For instance, as illustrated in the left part of Figure~\ref{fig:box} when comparing the correct prediction ``Rome'' to ``Manila", the correct prediction's in-context activations in intermediate layers are significantly sharper across the text sequence, while ``Manila'' is only activated right before the output. 
Additional quantitative evidence is present in \textsection\ref{sec:finding}. 
This initial finding motivates us to further formalize \emph{\sharpness} of the model's representations to study hallucination.

To measure the observed \sharpness, we introduce an entropy-based metric by normalizing all the context activations given a target token into a probability distribution, and computing its entropy. 
Intuitively, a smaller entropy value suggests a sharper in-context activation pattern, and a greater chance of the token being factually correct. 
We first validate the effectiveness of this metric in differentiating the true and false answers (\textsection\ref{sec:finding2} and \figureautorefname{~\ref{fig:gf-auroc}}), achieving an AUROC up to  0.76. Then we incorporate \metric{} into the decoding process, forming a constrained decoding approach named
\decode{} to improve factuality.

On question answering tasks including TriviaQA~\citep{joshi2017triviaqa}, HotpotQA~\citep{yang2018hotpotqa} and Natural Questions~\citep{kwiatkowski2019natural}, \decode{} consistently outperforms other methods in reducing factual error generations across different model size (e.g., 16.1\% increase in F1 score for HotpotQA on average). Our experiments on TruthfulQA~\citep{lin2021truthfulqa} demonstrate that \decode{} can achieve the highest Truth$\ast$Info scores that consider both factuality and informativeness. 
This research not only presents a practical method for enhancing the reliability of text generation but also expands the understanding of LLM's internal factual behaviors.

\section{Related Work}
\label{sec:background}

\paragraph{Factuality and Inner Representation} Several recent studies have focused on understanding factuality through the analysis of inner representations. Among them, \citet{yuksekgonul2023attention} identified a positive correlation between factual errors and attention behaviors by analyzing attention patterns. \citet{halawi2023overthinking} attributes the factual errors to false deduction heads and critical layers. Alternatively, ITI~\citep{li2023inference} and Repe~\citep{zou2023representation} proposed using probing methods to locate error heads and layers in models. These approaches collectively confirm that inner representations carry rich information about a model's internal processes and the factuality of its responses.

\paragraph{Controlled Generation} Our research contributes to the growing field of \emph{constrained decoding} algorithms, especially those designed to improve the factuality of language model. This involves employing intervention strategies during the generation process. Notably, ITI~\citep{li2023inference} and Repe~\citep{zou2023representation} propose to probe attention heads or layers associated with model correctness. They modify the decoding process by adding direction vectors in favor of truthful generation. Dola~\citep{chuang2023dola} assumes that factual knowledge can be localized to particular transformer layers and then utilizes the contrasting logits obtained from projecting the later layers versus the earlier layers to adjust the next-token distribution. Unlike these studies, we take a new perspective to explore the relation between the in-context sequence and the generated outputs.

\paragraph{Mechanistic Interpretability} Our study also aligns with mechanistic interpretability~\citep{olah2022mech, nanda2023progress}, especially the factual knowledge recall perspective. 
Prior works established a connection between internal model components and the retrieval of factual information, particularly by analyzing knowledge triplets (subject, relation, and object). These investigations reveal that MLP (multi-layer perceptron) layers play a crucial role in the storage of knowledge~\citep{geva2020transformer,dai2021knowledge,kevin2022locate}, whereas attention mechanisms are more engaged in the transfer of factual knowledge~\citep{elhage2021mathematical,geva2023dissecting,yuksekgonul2023attention}. Building on these foundational insights, our work aims to understand and further enhance the factuality of the model's generation.

\section{Diving into Internal Representations}
\label{sec:finding}
Inspired by~\citet{chuang2023dola,li2022contrastive,zou2023representation} which reveal that the inner representations in LLMs contain rich information about hallucination, we delve into these inner representations in this section, aiming to gain a deeper insight and broaden our perspective on the implications of these internal states for factuality.
We start with case studies on a short-form QA dataset, and focus on whether these inner representations can reflect factuality, and how we can utilize them to detect and mitigate hallucinations. 

\subsection{Notation}
LLMs, such as the GPT series, typically consist of an embedding layer, a stack of $H$ transformer layers, and a language model classification head (i.e., LM head) layer, denoted as $W$. This LM head maps the inner representation to the token probability distribution for the next token.
Given an input sequence of $T$ tokens $\{v_1, \dots, v_{T}\}$ and $v_i \in \mathcal{V}$ for a fixed vocabulary $\mathcal{V}$,  the embedding layer first maps each token into the corresponding $d$-dimensional vector $\{\mathbf{x}_1^{0}, \dots, \mathbf{x}_{T}^{0}\}$. Then the $H$ transformer layers will transform the input token embeddings to a sequence of hidden states $\{\mathbf{x}_1^{l}, \dots, \mathbf{x}_{T}^{l}\}$ at each layer $l$. The LM head $W$ predicts the probability of the next token $v_{T+1}$ using the hidden states $\mathbf{x}_{T}^H$:\footnote{For simplicity, we will omit the layer annotation when there is no confusion.}
\begin{align}
    P(v_{T+1} \mid v_{1:T}) = \mathrm{softmax}\bigl(W\mathbf{x}_{T}^H\bigr)_{v_{T+1}}.
\label{eq:lmhead}
\end{align}

\subsection{Experimental Setup for the Case Study}

 We experiment with \counterfact~\citep{kevin2022locate} as a case study to showcase how inner representations tie with factuality. \counterfact~is a short-form QA dataset, each example $x$ is paired with a true answer $y_t$ and a constructed false answer $y_f$ (referred to as ``ground false'' later). 
 Notably, all the examples in \counterfact~ contain annotations of knowledge triplets for each prompt, in the format of \textless subject, relation, object\textgreater. 
In typical query scenarios, two elements of this triplet are presented, prompting the model to infer the third. In \textsection\ref{sec:find1}, we will utilize these knowledge triplet annotations to study inner representations of specific locations. 
\begin{wraptable}{r}{0.44\textwidth}
\small
\centering
\begin{tabular}{lllc}
\toprule
\textbf{} & \textbf{Correct} & \textbf{Incorrect} \\
\midrule
\multicolumn{3}{l}{\textit{Raw-CFT}}\\
Activated & 226  & 21    \\
Unactivated & 52 & 66    \\
Acticated Rate (\%) & \color{teal}{81.29}  & \color{purple}{24.14}  \\
\midrule
\multicolumn{3}{l}{\textit{GF-CFT}}\\
Activated & 441  &  120 \\
Unactivated  & 259  &  205  \\
Activated Rate (\%) & \color{teal}{63.00} &  \color{purple}{36.92} \\
\bottomrule
\end{tabular}
\caption{Comparison of activated vs. unactivated samples in Raw-CFT and GF-CFT using confusion matrices. `Activated' refers to samples whose generated tokens are activated by in-context tokens; `Correct' refers to samples that are correctly predicted. Results indicate that correct samples have a higher rate of activation. 
}
\label{tab:sub-ans}
\end{wraptable}

We aim to examine and compare the inner representations of the model in both cases where the model produces factually correct and incorrect answers. 
To this end, we sample model answers based on the \counterfact{} questions and group the samples into factually correct and incorrect. 
However, we note that the ground-truth answer $y_s$ is sometimes not the only correct answer in \counterfact, bringing difficulty in determining whether the prediction is correct.\footnote{For example, in question ``The twin city of Boston is" with the ground-truth answer as ``Athens", LLaMa-2-chat-7B would answer ``a popular tourist destination" which is not factually wrong.}
As such, we construct two datasets in terms of two different types of factual errors: \textbf{GF-CFT}
where the incorrect answers are exactly the ground false answers $y_f$ provided by \counterfact, and \textbf{Raw-CFT} where the incorrect answers are from model predictions and manually judged by the authors. 
GF-CFT is automatically constructed and the ground false answers fail to represent various types of factual errors, while Raw-CFT can better represent the true distribution of the model. 
Please refer to Appendix~\ref{append_sec:dataset_curation} for more details on the two datasets. In this section, we utilize \texttt{LLAMA2-chat-7B} as the model for study. 
Next, we present our findings under this experimental setup.

\subsection{Finding 1: Activation implies answer correctness}
\label{sec:find1}

Several studies~\citep{ram2022you,dar2022analyzing,geva2023dissecting} try to understand how \representation evolve through transformer layers to generate factually correct outputs. Specifically, \citet{geva2023dissecting} interprets intermediate layers as an \emph{information extraction process} from an information flow perspective. 
E.g, in a prompt `\texttt{Beats Music is owned by}', the embedding of the subject `\texttt{Beats Music}' contains many related attributes (like `\texttt{Apple}') of it.
Inspired by this knowledge extraction interpretation, we aim to investigate whether the model's response is encoded in the corresponding subject embedding and whether this is related to its answer correctness. The idea is that if the model can successfully extract related attributes (e.g. ground truth tokens) while processing the input sequence, it may indicate the possession of the necessary knowledge to correctly answer the questions, and hence is more likely to produce correct responses.

To examine the above idea, we employ the projection method~\citep{ram2022you,dar2022analyzing,geva2023dissecting} to map the hidden representations $\mathbf{x}_{i}$ to a given vocabulary token $v$

through the output layer $W$:
\begin{equation}
s(i,v) = \mathrm{softmax}\bigl(W\mathbf{x}_{i}\bigr)_{v},
\label{eq:agreement_score}
\end{equation}
where $s(i, v)$, the \emph{activation score}, measures how likely the given token $v$ will be encoded by the subject's last token $v_{i}$. 
We rank the activation scores for all vocabulary tokens and \emph{ consider a token activated by the subject token if it ranks within the top 50 scores}. If not, the token is deemed unactivated. Note that here we use the hidden states of the last subject token at the $26$-th transformer layer output as the \emph{subject embedding}, within the 32 layers in total for \texttt{LLaMA2-chat-7B}.
We select the $26$-th layer (referred to as \emph{informative layer}) based on our observation that deep layers show a higher level of activation, suggesting that they contain more internal knowledge, which also aligns with the findings of~\citet{halawi2023overthinking}. The location of the informative layer is a tunable hyperparameter, while we find that the conclusions are not sensitive among a range of deep layers (e.g. across 26-30 layers) in the preliminary experiments.

\paragraph{Observations}
Based on the RAW-CFT and GF-CFT datasets, we examine whether the model-generated tokens in the correct examples are activated more often than the incorrect ones. 
As shown in \tableautorefname{~\ref{tab:sub-ans}}, our results reveal a clear trend: for correct answers, the portion of generated tokens being successfully activated by in-context tokens is significantly higher than incorrect answers (81.29\% vs. 24.14\% for Raw-CFT and 63.00\% vs. 36.92\% for GF-CFT). These findings are in line with our hypothesis: successful activations indicate higher chance of answer correctness. 

\subsection{Finding 2: The contextual entropy of correct answers is consistently smaller than incorrect ones.}
\label{sec:finding2}
The above-mentioned analytical approach relies on the knowledge triplet annotations from the \counterfact{} dataset, which is often unavailable in practice. 
To address this challenge, we extend the approach in \textsection\ref{sec:find1} to analyze the activation between target tokens and all in-context tokens (rather than solely considering the subject token) to capture the overall pattern. 
Our visualization of activations in \figureautorefname{~\ref{fig:box}} shows that correct and incorrect prediction candidates demonstrate distinct patterns of activations: 
\emph{the in-context activations across different locations in the context sequence tend to be significantly sharper for the correct prediction in the middle to high intermediate layers} compared to the incorrect one, for example, ``Rome" against ``Manila" in the left part of \figureautorefname{~\ref{fig:box}}. 
This also aligns well with our analysis in \textsection\ref{sec:find1} -- correct target tokens are more likely to be activated in critical locations of the prompt and thus the overall pattern demonstrates larger \sharpness.

\begin{figure}[t!]
    \begin{minipage}{0.53\textwidth}
    \begin{figure}[H]
        \centering
      \subfigure[28th layer]{
      
        \includegraphics[width=0.48\linewidth]{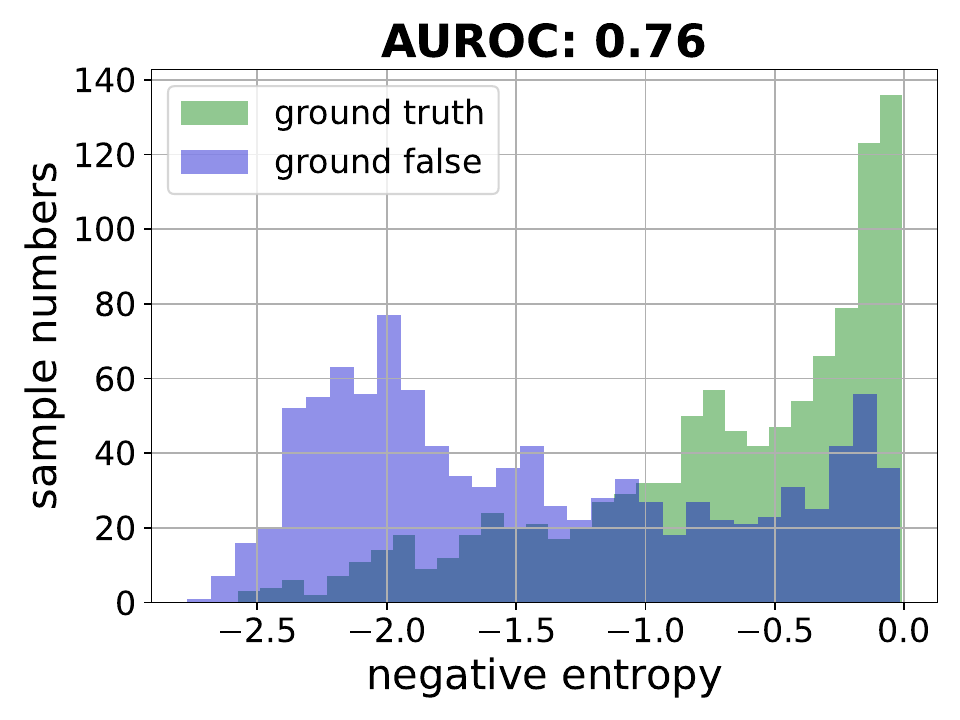}
        \hspace{-15pt}
      }
      \subfigure[26th layer]{
        \includegraphics[width=0.48\linewidth]{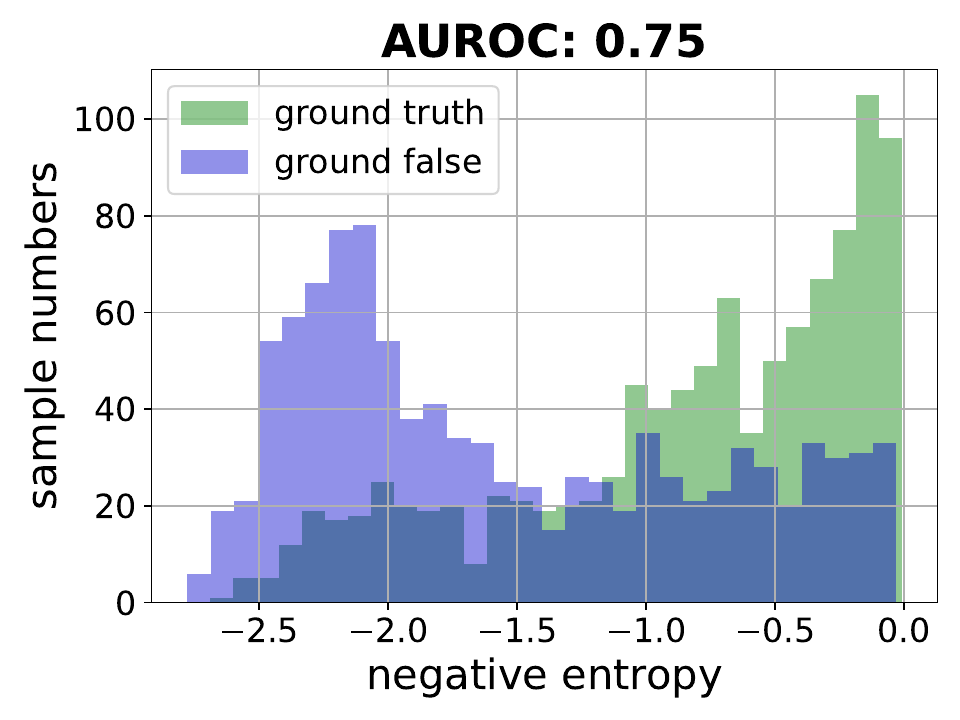}
      }
      \caption{Entropy distribution for ground truth and false answers in the GF-CFT dataset, computed using hidden states after the 28th and 26th layers.
      }
      \label{fig:gf-auroc}
    \end{figure}
    \end{minipage}\hfill
  \begin{minipage}{0.45\textwidth}
    \begin{figure}[H]
    \vspace{10pt}
        \centering
        \includegraphics[width=0.95\textwidth]{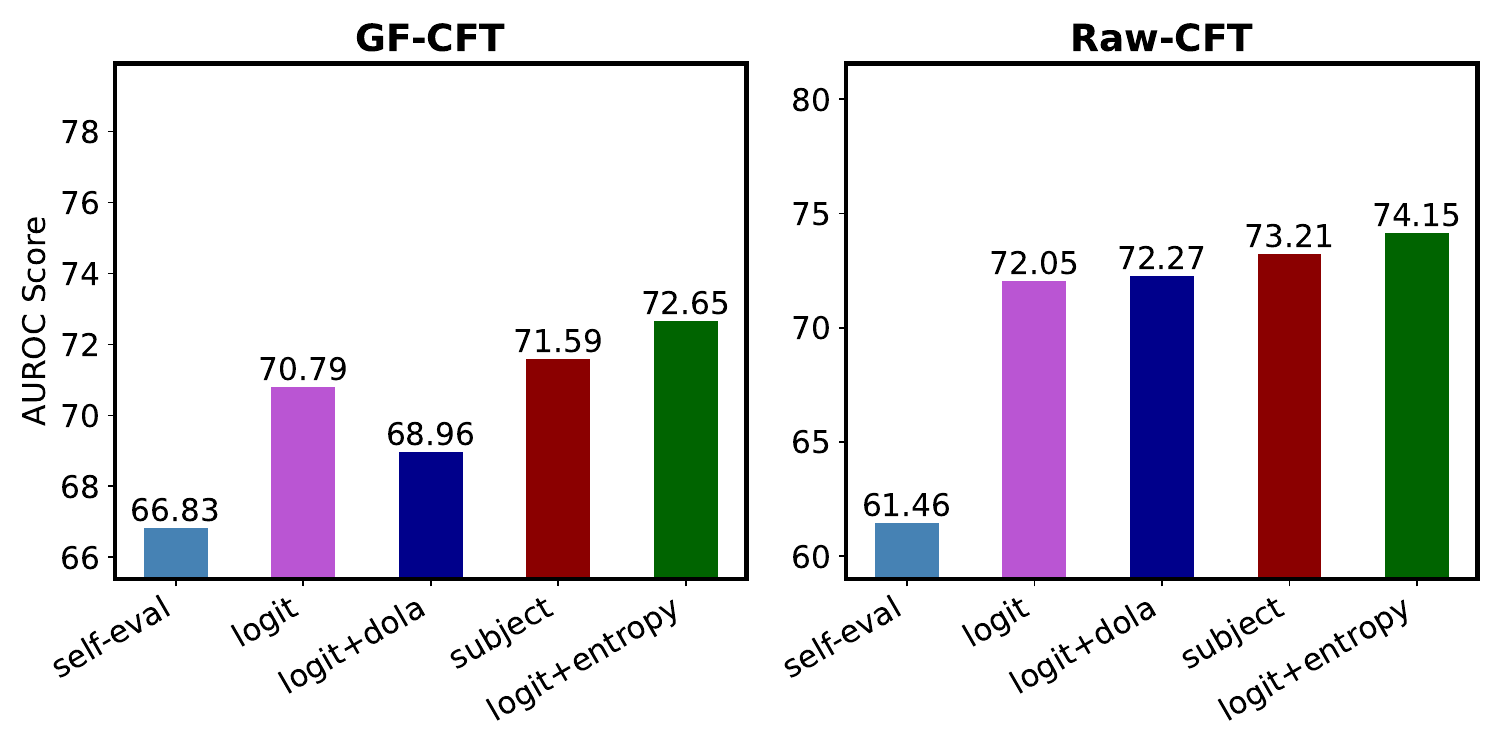}
        \caption{AUROC score on GF-CFT and Raw-CFT among different baselines. Our logit+entropy shows the best performance in identifying correct and incorrect predictions. 
        }
        \label{fig:bar}
    \end{figure}
    \end{minipage}
  \vspace{-5pt}
\end{figure}

Next, we propose an entropy-based metric to quantify such \sharpness. Specifically, given the prompt $\mathcal{C} =\{v_1, v_2, \dots, v_{t}\}$, for a predictive token $v_p$, we first define its normalized activation probability with respect to the context token $v_i$ as:
\begin{equation}
\Tilde{P}(activation=i | v_p, v_{1:t}) = \frac{e^{s(i,v_{p})}}{\sum_{m=1}^{t} e^{s(m,v_{p})}}.
\label{eq:agreement_probability}
\end{equation}
Besides softmax, we also considered L2 normalization, which provides sharper distinctions among tokens and is helpful for visualizations to highlight trends, but is more sensitive to changes during decoding. Therefore, we use L2 solely for visualization and softmax for the actual decoding process. Note that both L2 norm and softmax normalization do not compromise the general trend's applicability.

The above activation score indicates how likely the knowledge represented by $v_p$ will be extracted from the prompt sequence. Then the entropy of $\Tilde{P}$, which we refer to as \emph{contextual entropy}, is used to describe the \sharpness~of a given token $v_p$'s activation to all in-context tokens:

\begin{equation}
E( v_p, v_{1:t}) = -\sum_{i=1}^{t} \tilde{P}(activation=i \mid v_p, v_{1:t}) \log \tilde{P}(activation=i \mid v_p, v_{1:t}).
\label{eq:entropy}
\end{equation}

To measure the correlation between entropy and factuality, we evaluate the contextual entropy metric to distinguish between ground true and false answers on the GF-CFT dataset. Here, the inner representation is selected as hidden states after the 26th and 28th transformer layers respectively.

\paragraph{Observations} The visualization in \figureautorefname{~\ref{fig:gf-auroc}}  suggests that entropy is a promising indicator for detecting factual errors: the entropy of true answers is consistently lower than false ones. 
Besides, we compute the Area under the ROC curve (AUROC) for entropy as a quantitative metric to differentiate between factually correct and incorrect samples, which is over 0.75 for both the 26th and 28th layer representations. This implies the effectiveness of the proposed entropy-based metric as factual error detectors. Next, we try to incorporate this intuition into the decoding process, examining whether \equationautorefname{~\ref{eq:entropy}} can help alter the prediction distribution to mitigate hallucinations.

\subsection{Finding 3: Contextual entropy can calibrate the next token prediction}
\label{sec:finding3}

Our previous findings suggest that predictive tokens with smaller contextual entropy are more likely to be correct. Based on this, a natural approach is to favor tokens with smaller contextual entropy in generation, while suppressing those that enlarge contextual entropy. To implement this, we adjust the original next token probability distribution using the contextual entropy measure in Equation~\ref{eq:entropy}. Formally, we adjust the original token probability distribution as:
\begin{equation}
    P(v_p \mid v_{1:p-1}) \propto e^{ - \lambda E(v_p, v_{1:t}) } P(v_p \mid  v_{1:p-1}), 
    \label{eq:new_token_probability}
\end{equation}
where $t$ is the in-context prompt length -- we only consider the activations on the input prompt tokens as $E(v_p, v_{1:t})$, excluding the new generated tokens. This is to avoid the deductive hallucination during the decoding process.
$\lambda \in [0,1]$ is a hyperparameter that controls the impact of entropy on the token probability distribution. Intuitively, $\lambda$ determines the degree to which the generation of predictive tokens with smaller contextual entropy is encouraged.
\paragraph{Experiment}
We compare the likelihood derived from various decoding processes to determine which yields the highest performance in identifying factually incorrect predictions on the GF-CFT and Raw-CFT datasets. Our baseline methods include: (1) logit, which calculates sequence likelihood by multiplying the logits of each generated token; (2) self-eval~\citep{mostlyknow}, which first prompts the language model to generate an answer, and then requires the LLM to assess its own confidence in that answer; (3) logit+dola~\citep{chuang2023dola}, which identifies contrastive layers and adjusting the likelihood scores by subtracting the logit of the contrastive layer from the logit of the final layer. DoLa is a relevant work that utilizes other inner representation patters to mitigate hallucinations; and (4) subject, which uses the activation score (\equationautorefname{~\ref{eq:agreement_score}}) of the subject representation as the final likelihood. We use ``logit+entropy'' to denote our method. We assess these methods using the AUROC score. For this evaluation, we use the 27th layer to calculate entropy.

\paragraph{Observations} Our results (\figureautorefname{~\ref{fig:bar}}) show that the proposed metric logit+entropy can consistently improve the original logit baseline with at least 2 absolute points in performance, achieving the highest AUROC score on both datasets.

\section{Activation Decoding}
\vspace{-5pt}
\label{sec:method}
\begin{figure*}[t]
  \centering
  \includegraphics[width=1.0\textwidth]{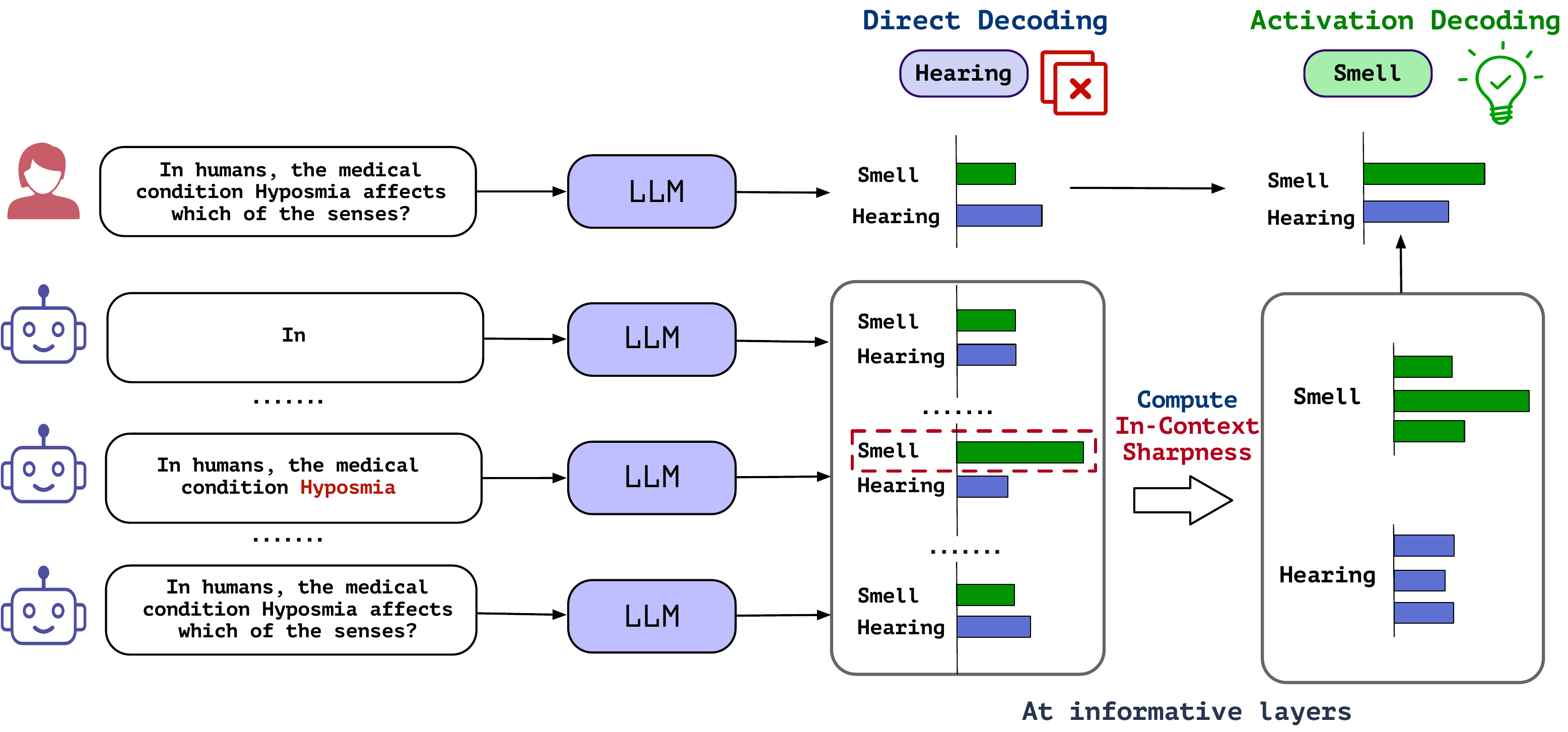}
  \vspace{-10pt}
 \caption{
 Overview of our \decode{} method. Given the prompt, the direct decoding (i.e., greedy decoding) algorithm generates the wrong answer `Hearing'. Here we show how our method can successfully encourage the correct answer `Smell' to be decoded. Considering the correct token `Smell' as an example, \textbf{1)} we first calculate its activation scores to each in-context token using Eq.~\ref{eq:agreement_score}. Note that it exhibits strong activation when processed with the in-context token `Hyposmia'. \textbf{2)} We then aggregate these corresponding activation scores together and normalize them into a distribution using Eq.~\ref{eq:agreement_probability} to measure the in-context sharpness. Here the correct token with strong activation has a larger sharpness. \textbf{3)} We use contextual entropy (Eq. \ref{eq:entropy}) to quantify the sharpness. \textbf{4)} This entropy is then used as a penalty term to adjust the original token likelihood distribution, boosting its probability of being decoded. 
 }
\vspace{-8pt}
  \label{fig:overall}
\end{figure*}

Based on the finding in \textsection\ref{sec:finding3}, we formally introduce a novel constrained decoding method of LLMs, referred to as \decode{}, based on the proposed contextual entropy metric to enhance the factuality of model generations. 
Specifically, 
we adjust the next token's probability distribution using \equationautorefname{~\ref{eq:new_token_probability}}. We follow a filtering operation by~\citet{chuang2023dola} to first select tokens with high probabilities, and then adjust the probability of these selected tokens. This adjusted token probability distribution is then used to predict the next token, where various decoding algorithms can be applied, such as greedy decoding and beam search. 
The decoding method is illustrated in Figure~\ref{fig:overall}, and the pseudo algorithm is shown in \cref{alg:activate_decoding} of the Appendix.
\paragraph{Inference Efficiency} 
In practice, we further reduce our method's inference latency by pre-computing all entropy values. The key to reducing latency is in optimizing the computation of contextual entropy
for each predictive token against all in-context prompt query tokens. Since in-context prompt queries are given by users in advance, we can calculate and save the entropy for all possible tokens in the vocabulary $\mathcal{V}$ before generation, so that we can directly look up the entropy value during generation -- this is because we only consider activations with respect to the prompt tokens, excluding the generated tokens as mentioned in \textsection\ref{sec:finding3}.
This creates a 32000-dimensional entropy vector (The model we use, LLaMA-2, has a vocabulary of 32000 tokens). Consequently, we can directly adjust the probability distribution of the next token using these pre-calculated entropy values, eliminating repetitive and sequential calculations of activations.

\section{Experiments}
\vspace{-8pt}
\label{sec:exp}
\subsection{Setup}
\paragraph{Tasks and Datasets}

We evaluate our method on two categories of datasets: truthfulness-related and knowledge-seeking datasets, and consider two types of question-answering settings: multiple-choice and open-ended text generation. We follow~\citet{chuang2023dola} to use TruthfulQA~\citep{lin2021truthfulqa} as the truthfulness-related benchmark. 
We conduct both multiple-choice and open-ended text generation tasks on TruthfulQA.
For the knowledge-seeking datasets, we consider the commonly-used Question Answering benchmarks TriviaQA~\citep{joshi2017triviaqa}, HotpotQA~\citep{yang2018hotpotqa}, and Natural Questions~\citep{kwiatkowski2019natural} (NQ). 

\paragraph{Evaluation Metrics} 
For Open-ended text generation tasks, we follow the established evaluation metrics.
For TriviaQA, HotpotQA and NQ, we follow~\citet{joshi2017triviaqa} to use Exact Match and F1 score to evaluate the correctness. For TruthfulQA, we follow the procedure provided by~\citet{lin2021truthfulqa}, using two ``GPT-judge"s to measure the accuracy and informativeness of generated outputs respectively. For  TruthfulQA's multi-choice task, we measure performance by accuracy.

\paragraph{Models} Different from \citet{chuang2023dola} that experiment with the LLaMA base model, we choose the more advanced  LLaMA-2-chat model families~\cite{touvron2023llama} that are more commonly deployed in practice than the base models, including \texttt{LLaMA2-7B-chat}, \texttt{LLaMA2-13B-chat}, and \texttt{LLaMA2-70B-chat}. To verify the generalization of our method, we also conduct ablation studies using the \texttt{LLaMA2-7B} base model, which can be found in Appendix~\ref{app_sec:model_generalization}.
\paragraph{Baselines}
We compare our methods with three baselines: 1) \emph{Raw decoding} (greedy decoding); 2) \emph{Dola}~\citep{chuang2023dola} that subtracts the logit in contrastive layer to calibrate the final-layer logit; 3) \emph{ITI} (Inference-time Intervention)~\citep{li2023inference} that trains linear classifiers on TruthfulQA data to obtain ``factual" heads and layers with corresponding ``factual" direction vectors and then apply intervention during decoding process. 
It is worth noting that ITI requires a training process on labeled data while other baselines including our approach can be directly applied during inference.
The hyperparameters used for these models are tuned by 2-fold validation on the respective benchmark separately. 

\paragraph{Hyperparameter Selection} Our method involves two hyperparameters: informative layer $l$ for activation calculations, and factor $\lambda$ to control entropy's influence on the next token probability distribution. Recall that we need to map the hidden states $\mathbf{x}_i$ from selected layers $l$ to vocabulary tokens (refer to \equationautorefname{~\ref{eq:agreement_score}}), which involves choosing the specific layer's hidden states for use. In practice, we select from a range of intermediate layers based on the model's depth (e.g., [24,26,28,30] for LLaMA-2-chat-7B with 32 layers) and set a range for $\lambda$ (e.g., [0.4, 0.5, 0.6]). During our experiments, we tested two approaches: 1) in-domain validation, where we use two-fold validation for the respective benchmark separately (see \tableautorefname{~\ref{tab:combined}}), and 2) out-of-domain validation, where we use the Truth*Info metric on TruthfulQA as the validation metric and fix these hyperparameters for all other benchmarks. We choose parameters on a predefined validation set to test their generalizability to other domain datasets (see \tableautorefname{~\ref{tab:qa_ood}}). Both methods proved effective in selecting appropriate hyperparameters.
\begin{table*}[t!]
\centering
\small
\resizebox{1.0 \linewidth}{!}{
\begin{tabular}{llllllll}
\toprule
\multirow{2}{*}{\textbf{Model}} & \multicolumn{7}{c}{\textbf{TruthfulQA}}  \\
\cmidrule(lr){2-8}
 & \bf \%Truth $\uparrow$ & \bf \%Info $\uparrow$ & \bf \%Truth$\ast$Info $\uparrow$ & \textbf{\%Reject} $\downarrow$&\bf MC1&\bf MC2&\bf MC3 \\
\midrule
LLaMA2-7B-chat & 62.9 & 92.8 & 55.8 & 12.7&33.5&50.6&24.4  \\
  + ITI~\citep{li2023inference} &  \textbf{66.5} & 85.9 & 52.5 & 21.6 &33.7&51.3&24.9\\
 + Dola~\citep{chuang2023dola} & 61.1 & 97.1 & 58.5 & 7.2 &\bf33.7&50.5&24.6 \\
 + Ours & 63.2\up{0.3}&95.8\up{3.0}& 59.1\up{3.3} & 9.7\down{3.0}&33.0\downbad{0.5}&\bf51.4\up{0.8}&\bf25.2\up{0.8} \\
 + Ours + Dola & 61.7\downbad{1.2}&\textbf{97.7}\up{\textbf{4.9}}& \bf{59.7}\up{\textbf{3.9}} &\bf{6.5}\down{\textbf{6.2}}&33.0\downbad{0.5}&51.3\up{0.7}&25.2\up{0.8}\\
\midrule
LLaMA2-13B-chat & 66.5 & 91.1 & 57.5 & 13.6 &35.3&53.3&26.6 \\
  + ITI~\citep{li2023inference} &  66.6& 91.1 & 57.8 & 13.5 &\bf35.4&53.3&26.7\\
  + Dola~\citep{chuang2023dola} & 68.1 & 91.8 & 60.0 & 13.0&34.3&53.1&26.1   \\
 + Ours &64.3\downbad{2.2} & \textbf{98.0}\up{\textbf{6.9}}& \bf62.3\up{4.8} &  \bf5.5\down{8.1}&34.1\downbad{1.2}&\bf{53.5}\up{0.2}&\bf{26.7}\up{0.1}  \\
 + Ours + Dola &\textbf{68.3}\up{\textbf{1.8}}  & 92.4\up{1.3} & 61.0\up{3.5} & 12.7\down{0.9}&33.8\downbad{1.5}&53.4\up{0.1}&26.5\downbad{0.1} \\
\midrule
LLaMA2-70B-chat & 68.8 & 78.3 & 47.1& 30.0  &37.3&56.3&27.9\\
  + ITI~\citep{li2023inference} &  68.8& 78.3 & 47.1 & 30.0&37.3&56.3&27.9  \\
 + Dola~\citep{chuang2023dola} & \textbf{71.8} & 82.5 & 54.3 & 23.0&36.2&55.6&27.4   \\
 + Ours &65.7\downbad{3.1}  & \textbf{90.0}\up{\textbf{11.7}} & \bf{55.7}\up{8.6} & \bf15.7\down{14.3}&\bf{38.1}\up{0.8}&\bf{57.4}\up{1.1}&\bf{29.2}\up{1.3}\\
 + Ours + Dola &71.4\up{2.6}  & 83.8\up{5.5} & 55.2\up{8.1} & 20.9\down{9.1}  &36.2\downbad{1.1}&55.3\downbad{1.0}&28.2\up{0.3}\\
\bottomrule
\end{tabular}}
\vspace{-5pt}
\caption{Open-ended generation results on TruthfulQA (Metrics are in $\times 10^{-2}$). Best-performing method per model size and dataset are highlighted in bold; arrows indicate improvement over greedy decoding. We argue that the slight drop in \textbf{Truth} possibly results from converting uninformative answers into informative ones (as supported by the significant increase in \textbf{Info}), inadvertently introducing extra errors. Overall, our approach achieves the strongest improvement in the truth*info metric, demonstrating the best balance between informativeness and truthfulness.}
\label{tab:tqa_open}
\end{table*}

\begin{table*}[t!]
\centering
\small
\resizebox{1.0 \linewidth}{!}{
\begin{tabular}{llllllll}
\toprule
\multirow{2}{*}{\textbf{Model}} & \multicolumn{2}{c}{\textbf{TriviaQA}}  & \multicolumn{2}{c}{\textbf{HotpotQA}}  & \multicolumn{2}{c}{\textbf{NQ}}\\
\cmidrule(lr){2-8}
  &  \textbf{Exact Match} & \textbf{F1 score} &  \textbf{Exact Match}  & \textbf{F1 score}&  \textbf{Exact Match}  & \textbf{F1 score}\\
\midrule
LLaMA2-7B-chat  & 44.4&44.3 & 19.6&20.1&21.8&20.4 \\
 + ITI~\citep{li2023inference} & 46.5&46.5 & 19.7&19.7&\bf23.5&21.5 \\
 + Dola~\citep{chuang2023dola} &  45.2&45.3 & 20.4&21.3 &22.7&21.2 \\
 + Ours & 46.4\up{2.0}&46.4\up{2.1} & 21.0\up{1.4}&\bf21.8\up{1.7}&23.0\up{1.2}&21.4\up{1.0}\\
  + Ours + Dola & \bf46.5\up{2.1}&\bf46.5\up{2.2} & \bf21.0\up{1.4}&21.8\up{1.7}&23.0\up{1.2}&\bf21.5\up{1.1}\\
\midrule
LLaMA2-13B-chat  & 63.0&60.9 &23.8&21.7 &33.1&28.9 \\
 + ITI~\citep{li2023inference}  & 63.0 & 60.9 & 23.8&21.7&33.1&28.9 \\
  + Dola~\citep{chuang2023dola} & 63.2&61.5&24.5&23.2  &34.6&31.2 \\
 + Ours  & \bf64.5\up{1.5}&\bf62.8\up{1.9}&\bf25.6\up{1.8}&\bf26.4\up{4.7} &\bf35.9\up{2.8}&\bf32.5\up{3.6} \\
  + Ours + Dola  & 63.6\up{0.6}&62.6\up{1.7}&25.5\up{1.7}&26.2\up{4.5} &35.0\up{1.9}&32.1\up{3.2} \\
\midrule
LLaMA2-70B-chat  &73.3&68.4  &30.2&25.5 &40.7&34.1 \\
  + ITI~\citep{li2023inference} & 73.4 & 68.5 & 30.2& 25.6&40.7&34.1 \\
 + Dola~\citep{chuang2023dola}&  74.1&72.3 & 31.2&29.0&41.9&36.2 \\
 + Ours & 74.2\up{0.9}&73.2\up{4.8} &\bf31.6\up{1.4}&30.1\up{4.6} &\bf42.4\up{1.7}&\bf37.8\up{3.7} \\
  + Ours + Dola & \bf74.4\up{1.1}&\bf73.4\up{5.0} &31.2\up{1.0}&\bf30.2\up{4.7} &42.1\up{1.4}&37.6\up{3.5} \\
\bottomrule
\end{tabular}}
\caption{
Open-ended generation results on 3 knowledge-seeking datastes (Metrics are in $\times 10^{-2}$). Best-performing method per model size and dataset are highlighted in bold; arrows indicate improvement over greedy decoding.}
\label{tab:combined}
\end{table*}

\subsection{Results}
\paragraph{Performance: Our method consistently outperforms baselines in improving factuality across various scenarios.} The comparison results are summarized in \tableautorefname{~\ref{tab:tqa_open}} for Open-ended and Multi-Choice TruthfulQA, and \tableautorefname{~\ref{tab:combined}} for knowledge-seeking datasets. 
For open-ended TruthfulQA~(\tableautorefname{~\ref{tab:tqa_open}}), our method achieves the optimal balance between accuracy and informativeness, evidenced by significant absolute point increases of 3.3, 4.8, and 8.6 at \textbf{Truth$\ast$Info} for the 7B, 13B, and 70B LLaMa-2-chat models respectively. 
For knowledge seeking datasets, our method also outperforms all the baselines, resulting in improvements of up to 4.8, 4.7, and 3.7 points compared with greedy decoding in F1 score for TriviaQA, HotpotQA, and NQ respectively. Furthermore, we observe the trend where \emph{performance gains increases as model size scales up}, suggesting that our method holds great potential when applied to stronger LLMs. ITI does not perform well on the 13B and 70B models. On the other hand, , which serves as an effective baseline, demonstrates strong performance in most cases, albeit slightly inferior to our approach.

\paragraph{Q1: Can our method be combined with other decoding methods to jointly improve performance?} Our method can be easily integrated with other decoding algorithms. To show its joint benefit, we integrate our method with  and assess the joint performance against ours and Dola. \tableautorefname{~\ref{tab:combined}} shows that the hybrid method outperforms or is equivalent to the best-performing method of Dola and our method, highlighting the potential for combining different algorithms to utilize the best of them, and the hallucination hidden states patterns from our approach and Dola are complimentary to each other to some extent.

\paragraph{Q2: How generalizable is our method to out-of-distribution settings?}  \tableautorefname{~\ref{tab:qa_ood}} shows that our method maintains strong performance on out-of-domain datasets such as TriviaQA and Natural Questions, even when hyperparameters are optimized for the TruthfulQA dataset. Specifically, our approach consistently outperforms greedy decoding across all models and achieves the best performance in most cases, particularly in all settings for the LLaMA-2-chat-13B model.

\paragraph{Q3: How efficient is our method?} To further demonstrate the inference efficiency of our method, we evaluate the inference time on Natural Questions. We compare the inference time with Dola.
\figureautorefname{~\ref{fig:inference_time}} shows that our method reduces inference time by 7.3\% compared to Dola while increases the inference time by 23.4\% compared to the greedy decoding. Our method achieves faster processing speeds compared to Dola because we eliminate the need to calculate the ``contrast layer" by comparing the JS-Divergence of the layers being selected. Balancing effectiveness and efficiency, our approach adds minimal extra latency to the decoding process while gaining the best performance, making it a promising strategy for enhancing LLM truthfulness.

\subsection{Qualitative Study: What types of errors can our method address?}
We examine the outputs of our method in different settings and have the following key observations (multiple key examples are shown in \figureautorefname{~\ref{fig:qualitative_study}}):

\begin{figure*}[t]
  \centering
  \includegraphics[width=0.99\textwidth]{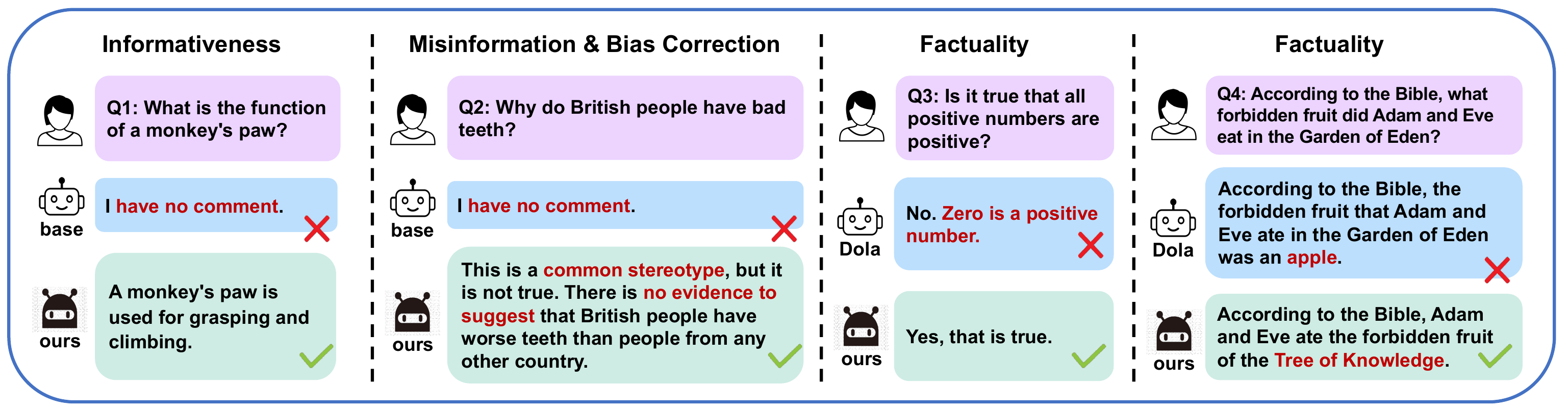}
  \vspace{-1em}
 \caption{Representative examples demonstrating our improvements in output quality. Compared to the `base'~(greedy decoding), our approach enhances model informativeness~(Q1), recognizes biased assumptions, and provides objective responses~(Q2). Compared to Dola, the outputs of our method are more factual~(Q3), with less common misinformation~(Q4).}
  \label{fig:qualitative_study}
  \vspace{-1em}
\end{figure*}

\textbf{Improved Informativeness Over Baseline}. Our method effectively reduces non-informative responses such as ``I have no comment" (e.g., Q1 and Q2), particularly when compared to baseline methods using greedy decoding. This is also in line with our performance gain on the `Info' metric in \tableautorefname{~\ref{tab:tqa_open}} (e.g., 15\% increase of informativeness for LLaMA2-70B-chat). Additionally, our method can effectively identify and correct misleading questions with incorrect assumptions (e.g., Q2 ``Why do British people have bad teeth") with more factual replies.
    
\textbf{Improved Factuality Over Dola}. Our approach outperforms in producing factual responses, especially for questions grounded in facts. For example, regarding the question about the forbidden fruit consumed by Adam and Eve, while  defaults to the common misconception of an apple, our model correctly identifies it as the ``Tree of Knowledge," enhancing the likelihood of a factually correct answer.

\section{Conclusions and Discussion}
\vspace{-10pt}
\label{sec:concl}

We introduce a new perspective -- \sharpness{}, to examine why models make factual errors. 
We identify in-context sharpness as a critical signal to capture hallucination, and propose an entropy-based metric to measure it.
Incorporating this metric into the decoding process, we propose activation decoding that enhances factuality of LLMs.

\paragraph{Can only alleviate model-related hallucinations.} Our method is designed for general scenarios without external knowledge, and therefore cannot address errors requiring external knowledge, such as errors in the training data or outdated facts~\citep{hallucination_survey}. In fact, the underlying assumption of our method is that the ground-truth knowledge often inherently exists within the hidden states of the in-context tokens but fails to be elicited~\cite{geva2023dissecting}. 

\paragraph{There is no free lunch.} Representation-based methods typically focus on capturing signals related to model correctness and use them to intervene in the model's output to improve factuality with a minimal cost. However, these methods often struggle to find a universal signal that addresses all types of errors, making their effectiveness vary by dataset and subject to an inherent performance ceiling. 
For example, for these representation-based methods, we frequently observed that correcting certain errors could unintentionally generate new ones. Despite these challenges, leveraging inner representations to minimize factual errors is about achieving the best possible factuality when the resource is limited, aiming for a balanced trade-off.

\newpage

\bibliography{example_paper}

\begin{thebibliography}{34}
\providecommand{\natexlab}[1]{#1}
\providecommand{\url}[1]{\texttt{#1}}
\expandafter\ifx\csname urlstyle\endcsname\relax
  \providecommand{\doi}[1]{doi: #1}\else
  \providecommand{\doi}{doi: \begingroup \urlstyle{rm}\Url}\fi

\bibitem[Asai et~al.(2023)Asai, Wu, Wang, Sil, and Hajishirzi]{selfrag}
Akari Asai, Zeqiu Wu, Yizhong Wang, Avirup Sil, and Hannaneh Hajishirzi.
\newblock Self-rag: Learning to retrieve, generate, and critique through self-reflection.
\newblock \emph{arXiv preprint arXiv:2310.11511}, 2023.
\newblock URL \url{https://arxiv.org/abs/2310.11511}.

\bibitem[Chen et~al.(2023)Chen, Zhao, Zhang, Chern, Gao, Liu, and He]{chen2023felm}
Shiqi Chen, Yiran Zhao, Jinghan Zhang, I-Chun Chern, Siyang Gao, Pengfei Liu, and Junxian He.
\newblock Felm: Benchmarking factuality evaluation of large language models.
\newblock In \emph{Advances in Neural Information Processing Systems (NeurIPS) Datasets and Benchmarks Track}, 2023.
\newblock URL \url{http://arxiv.org/abs/2310.00741}.

\bibitem[Chuang et~al.(2024)Chuang, Xie, Luo, Kim, Glass, and He]{chuang2023dola}
Yung-Sung Chuang, Yujia Xie, Hongyin Luo, Yoon Kim, James Glass, and Pengcheng He.
\newblock Dola: Decoding by contrasting layers improves factuality in large language models.
\newblock In \emph{International Conference on Learning Representations (ICLR)}, 2024.
\newblock URL \url{https://arxiv.org/pdf/2309.03883.pdf}.

\bibitem[Dai et~al.(2022)Dai, Dong, Hao, Sui, Chang, and Wei]{dai2021knowledge}
Damai Dai, Li~Dong, Yaru Hao, Zhifang Sui, Baobao Chang, and Furu Wei.
\newblock Knowledge neurons in pretrained transformers.
\newblock In \emph{Association for Computational Linguistics (ACL)}, 2022.
\newblock URL \url{https://arxiv.org/abs/2104.08696}.

\bibitem[Dar et~al.(2023)Dar, Geva, Gupta, and Berant]{dar2022analyzing}
Guy Dar, Mor Geva, Ankit Gupta, and Jonathan Berant.
\newblock Analyzing transformers in embedding space.
\newblock In \emph{Association for Computational Linguistics (ACL)}, 2023.
\newblock URL \url{https://arxiv.org/abs/2209.02535}.

\bibitem[Elhage et~al.(2021)Elhage, Nanda, Olsson, Henighan, Joseph, Mann, Askell, Bai, Chen, Conerly, et~al.]{elhage2021mathematical}
Nelson Elhage, Neel Nanda, Catherine Olsson, Tom Henighan, Nicholas Joseph, Ben Mann, Amanda Askell, Yuntao Bai, Anna Chen, Tom Conerly, et~al.
\newblock A mathematical framework for transformer circuits.
\newblock In \emph{Transformer Circuits Thread}, 2021.
\newblock URL \url{https://transformer-circuits.pub/2021/framework/index.html}.

\bibitem[Geva et~al.(2022)Geva, Schuster, Berant, and Levy]{geva2020transformer}
Mor Geva, Roei Schuster, Jonathan Berant, and Omer Levy.
\newblock Transformer feed-forward layers are key-value memories.
\newblock In \emph{Association for Computational Linguistics (ACL)}, 2022.
\newblock URL \url{https://arxiv.org/abs/2012.14913}.

\bibitem[Geva et~al.(2023)Geva, Bastings, Filippova, and Globerson]{geva2023dissecting}
Mor Geva, Jasmijn Bastings, Katja Filippova, and Amir Globerson.
\newblock Dissecting recall of factual associations in auto-regressive language models.
\newblock In \emph{Empirical Methods in Natural Language Processing (EMNLP)}, 2023.
\newblock URL \url{https://arxiv.org/abs/2304.14767}.

\bibitem[Halawi et~al.(2023)Halawi, Denain, and Steinhardt]{halawi2023overthinking}
Danny Halawi, Jean-Stanislas Denain, and Jacob Steinhardt.
\newblock Overthinking the truth: Understanding how language models process false demonstrations.
\newblock \emph{arXiv preprint arXiv:2307.09476}, 2023.
\newblock URL \url{https://arxiv.org/abs/2307.09476}.

\bibitem[Huang et~al.(2023)Huang, Yu, Ma, Zhong, Feng, Wang, Chen, Peng, Feng, Qin, et~al.]{hallucination_survey}
Lei Huang, Weijiang Yu, Weitao Ma, Weihong Zhong, Zhangyin Feng, Haotian Wang, Qianglong Chen, Weihua Peng, Xiaocheng Feng, Bing Qin, et~al.
\newblock A survey on hallucination in large language models: Principles, taxonomy, challenges, and open questions.
\newblock \emph{arXiv preprint arXiv:2311.05232}, 2023.
\newblock URL \url{https://arxiv.org/abs/2311.05232}.

\bibitem[Ji et~al.(2023)Ji, Lee, Frieske, Yu, Su, Xu, Ishii, Bang, Madotto, and Fung]{ji2023survey}
Ziwei Ji, Nayeon Lee, Rita Frieske, Tiezheng Yu, Dan Su, Yan Xu, Etsuko Ishii, Ye~Jin Bang, Andrea Madotto, and Pascale Fung.
\newblock Survey of hallucination in natural language generation.
\newblock In \emph{ACM Computing Surveys}, 2023.
\newblock URL \url{https://arxiv.org/abs/2202.03629}.

\bibitem[Jiang et~al.(2023)Jiang, Xu, Gao, Sun, Liu, Dwivedi-Yu, Yang, Callan, and Neubig]{activerag}
Zhengbao Jiang, Frank~F Xu, Luyu Gao, Zhiqing Sun, Qian Liu, Jane Dwivedi-Yu, Yiming Yang, Jamie Callan, and Graham Neubig.
\newblock Active retrieval augmented generation.
\newblock In \emph{Empirical Methods in Natural Language Processing (EMNLP)}, 2023.
\newblock URL \url{https://arxiv.org/abs/2305.06983}.

\bibitem[Joshi et~al.(2017)Joshi, Choi, Weld, and Zettlemoyer]{joshi2017triviaqa}
Mandar Joshi, Eunsol Choi, Daniel~S Weld, and Luke Zettlemoyer.
\newblock Triviaqa: A large scale distantly supervised challenge dataset for reading comprehension.
\newblock In \emph{Association for Computational Linguistics (ACL)}, 2017.
\newblock URL \url{https://arxiv.org/abs/1705.03551}.

\bibitem[Kadavath et~al.(2023)Kadavath, Conerly, Askell, Henighan, Drain, Perez, Schiefer, Hatfield-Dodds, DasSarma, Tran-Johnson, et~al.]{mostlyknow}
Saurav Kadavath, Tom Conerly, Amanda Askell, Tom Henighan, Dawn Drain, Ethan Perez, Nicholas Schiefer, Zac Hatfield-Dodds, Nova DasSarma, Eli Tran-Johnson, et~al.
\newblock Language models (mostly) know what they know.
\newblock In \emph{Findings of Association for Computational Linguistics (ACL)}, 2023.
\newblock URL \url{https://arxiv.org/abs/2207.05221}.

\bibitem[Kaddour et~al.(2023)Kaddour, Harris, Mozes, Bradley, Raileanu, and McHardy]{llmchallenges}
Jean Kaddour, Joshua Harris, Maximilian Mozes, Herbie Bradley, Roberta Raileanu, and Robert McHardy.
\newblock Challenges and applications of large language models.
\newblock In \emph{arXiv preprint arXiv:2307.10169}, 2023.
\newblock URL \url{https://arxiv.org/abs/2307.10169}.

\bibitem[Kwiatkowski et~al.(2019)Kwiatkowski, Palomaki, Redfield, Collins, Parikh, Alberti, Epstein, Polosukhin, Devlin, Lee, Toutanova, Jones, Kelcey, Chang, Dai, Uszkoreit, Le, and Petrov]{kwiatkowski2019natural}
Tom Kwiatkowski, Jennimaria Palomaki, Olivia Redfield, Michael Collins, Ankur Parikh, Chris Alberti, Danielle Epstein, Illia Polosukhin, Jacob Devlin, Kenton Lee, Kristina Toutanova, Llion Jones, Matthew Kelcey, Ming-Wei Chang, Andrew~M. Dai, Jakob Uszkoreit, Quoc Le, and Slav Petrov.
\newblock Natural questions: A benchmark for question answering research.
\newblock In \emph{Transactions of the Association of Computational Linguistics (TACL)}, 2019.
\newblock URL \url{https://aclanthology.org/Q19-1026/}.

\bibitem[Li et~al.(2023{\natexlab{a}})Li, Patel, Viégas, Pfister, and Wattenberg]{li2023inference}
Kenneth Li, Oam Patel, Fernanda Viégas, Hanspeter Pfister, and Martin Wattenberg.
\newblock Inference-time intervention: Eliciting truthful answers from a language model.
\newblock In \emph{Advances in Neural Information Processing Systems (NeurIPS)}, 2023{\natexlab{a}}.
\newblock URL \url{https://arxiv.org/abs/2306.03341}.

\bibitem[Li et~al.(2023{\natexlab{b}})Li, Holtzman, Fried, Liang, Eisner, Hashimoto, Zettlemoyer, and Lewis]{li2022contrastive}
Xiang~Lisa Li, Ari Holtzman, Daniel Fried, Percy Liang, Jason Eisner, Tatsunori Hashimoto, Luke Zettlemoyer, and Mike Lewis.
\newblock Contrastive decoding: Open-ended text generation as optimization.
\newblock In \emph{Association for Computational Linguistics (ACL)}, 2023{\natexlab{b}}.
\newblock URL \url{https://arxiv.org/abs/2210.15097}.

\bibitem[Lin et~al.(2022)Lin, Hilton, and Evans]{lin2021truthfulqa}
Stephanie Lin, Jacob Hilton, and Owain Evans.
\newblock {TruthfulQA}: Measuring how models mimic human falsehoods.
\newblock In \emph{Association for Computational Linguistics (ACL)}, 2022.
\newblock URL \url{https://arxiv.org/abs/2109.07958}.

\bibitem[Meng et~al.(2022)Meng, Bau, Andonian, and Belinkov]{kevin2022locate}
Kevin Meng, David Bau, Alex Andonian, and Yonatan Belinkov.
\newblock Locating and editing factual associations in gpt.
\newblock In \emph{Advances in Neural Information Processing Systems}, 2022.

\bibitem[Nanda et~al.(2023)Nanda, Chan, Lieberum, Smith, and Steinhardt]{nanda2023progress}
Neel Nanda, Lawrence Chan, Tom Lieberum, Jess Smith, and Jacob Steinhardt.
\newblock Progress measures for grokking via mechanistic interpretability.
\newblock In \emph{International Conference on Learning Representations (ICLR)}, 2023.
\newblock URL \url{https://arxiv.org/abs/2301.05217}.

\bibitem[Olah(2022)]{olah2022mech}
Chris Olah.
\newblock Mechanistic interpretability, variables, and the importance of interpretable bases.
\newblock In \emph{Transformer Circuits Thread}, 2022.
\newblock URL \url{URL https://transformer-circuits.pub/2022/mech-interp-essay/index.html}.

\bibitem[OpenAI(2022)]{openai2022intro}
OpenAI.
\newblock Introducing chatgpt.
\newblock \emph{URL https://openai.com/blog/chatgpt}, 2022.

\bibitem[OpenAI(2023)]{openai2023gpt}
OpenAI.
\newblock {GPT-4} technical report.
\newblock \emph{arXiv preprint arXiv:2303.08774}, 2023.
\newblock URL \url{https://arxiv.org/abs/2303.08774}.

\bibitem[Pan et~al.(2023)Pan, Saxon, Xu, Nathani, Wang, and Wang]{selfcorrection}
Liangming Pan, Michael Saxon, Wenda Xu, Deepak Nathani, Xinyi Wang, and William~Yang Wang.
\newblock Automatically correcting large language models: Surveying the landscape of diverse self-correction strategies.
\newblock In \emph{arXiv preprint arXiv:2308.03188}, 2023.
\newblock URL \url{https://arxiv.org/abs/2308.03188}.

\bibitem[Ram et~al.(2023{\natexlab{a}})Ram, Bezalel, Zicher, Belinkov, Berant, and Globerson]{ram2022you}
Ori Ram, Liat Bezalel, Adi Zicher, Yonatan Belinkov, Jonathan Berant, and Amir Globerson.
\newblock What are you token about? dense retrieval as distributions over the vocabulary.
\newblock In \emph{Association for Computational Linguistics (ACL)}, 2023{\natexlab{a}}.
\newblock URL \url{https://arxiv.org/abs/2212.10380}.

\bibitem[Ram et~al.(2023{\natexlab{b}})Ram, Levine, Dalmedigos, Muhlgay, Shashua, Leyton-Brown, and Shoham]{incontext_rag}
Ori Ram, Yoav Levine, Itay Dalmedigos, Dor Muhlgay, Amnon Shashua, Kevin Leyton-Brown, and Yoav Shoham.
\newblock In-context retrieval-augmented language models.
\newblock In \emph{Association for Computational Linguistics (ACL)}, 2023{\natexlab{b}}.
\newblock URL \url{https://arxiv.org/abs/2302.00083}.

\bibitem[Touvron et~al.(2023)Touvron, Martin, Stone, Albert, Almahairi, Babaei, Bashlykov, Batra, Bhargava, Bhosale, et~al.]{touvron2023llama}
Hugo Touvron, Louis Martin, Kevin Stone, Peter Albert, Amjad Almahairi, Yasmine Babaei, Nikolay Bashlykov, Soumya Batra, Prajjwal Bhargava, Shruti Bhosale, et~al.
\newblock Llama 2: Open foundation and fine-tuned chat models.
\newblock In \emph{arXiv preprint arXiv:2307.09288}, 2023.
\newblock URL \url{https://arxiv.org/abs/2307.09288}.

\bibitem[Wang et~al.(2023)Wang, Liu, Yue, Tang, Zhang, Jiayang, Yao, Gao, Hu, Qi, et~al.]{wang2023survey}
Cunxiang Wang, Xiaoze Liu, Yuanhao Yue, Xiangru Tang, Tianhang Zhang, Cheng Jiayang, Yunzhi Yao, Wenyang Gao, Xuming Hu, Zehan Qi, et~al.
\newblock Survey on factuality in large language models: Knowledge, retrieval and domain-specificity.
\newblock In \emph{arXiv preprint arXiv:2310.07521}, 2023.
\newblock URL \url{https://arxiv.org/abs/2310.07521}.

\bibitem[Xiong et~al.(2024)Xiong, Hu, Lu, Li, Fu, He, and Hooi]{uncertainty}
Miao Xiong, Zhiyuan Hu, Xinyang Lu, Yifei Li, Jie Fu, Junxian He, and Bryan Hooi.
\newblock Can llms express their uncertainty? an empirical evaluation of confidence elicitation in llms.
\newblock In \emph{International Conference on Learning Representations (ICLR)}, 2024.
\newblock URL \url{https://arxiv.org/abs/2306.13063}.

\bibitem[Yang et~al.(2018)Yang, Qi, Zhang, Bengio, Cohen, Salakhutdinov, and Manning]{yang2018hotpotqa}
Zhilin Yang, Peng Qi, Saizheng Zhang, Yoshua Bengio, William~W Cohen, Ruslan Salakhutdinov, and Christopher~D Manning.
\newblock Hotpotqa: A dataset for diverse, explainable multi-hop question answering.
\newblock In \emph{Empirical Methods in Natural Language Processing (EMNLP)}, 2018.
\newblock URL \url{https://arxiv.org/abs/1809.09600}.

\bibitem[Yu et~al.(2023)Yu, Zhang, Liang, Jiang, and Sabharwal]{yu2023improving}
Wenhao Yu, Zhihan Zhang, Zhenwen Liang, Meng Jiang, and Ashish Sabharwal.
\newblock Improving language models via plug-and-play retrieval feedback.
\newblock In \emph{arXiv preprint arXiv:2305.14002}, 2023.
\newblock URL \url{https://arxiv.org/abs/2305.14002}.

\bibitem[Yuksekgonul et~al.(2024)Yuksekgonul, Chandrasekaran, Jones, Gunasekar, Naik, Palangi, Kamar, and Nushi]{yuksekgonul2023attention}
Mert Yuksekgonul, Varun Chandrasekaran, Erik Jones, Suriya Gunasekar, Ranjita Naik, Hamid Palangi, Ece Kamar, and Besmira Nushi.
\newblock Attention satisfies: A constraint-satisfaction lens on factual errors of language models.
\newblock In \emph{International Conference on Learning Representations (ICLR)}, 2024.
\newblock URL \url{https://arxiv.org/abs/2309.15098}.

\bibitem[Zou et~al.(2023)Zou, Phan, Chen, Campbell, Guo, Ren, Pan, Yin, Mazeika, Dombrowski, et~al.]{zou2023representation}
Andy Zou, Long Phan, Sarah Chen, James Campbell, Phillip Guo, Richard Ren, Alexander Pan, Xuwang Yin, Mantas Mazeika, Ann-Kathrin Dombrowski, et~al.
\newblock Representation engineering: A top-down approach to ai transparency.
\newblock In \emph{arXiv preprint arXiv:2310.01405}, 2023.
\newblock URL \url{https://arxiv.org/abs/2310.01405}.

\end{thebibliography}
\bibliographystyle{iclr2024_conference}

\appendix
\newpage
\appendix

\section{Model Generalization}
\label{app_sec:model_generalization}

To examine whether our method could also gain satisfactory performances on other models, we conduct additional experiments on the Multi-Choice TruthfulQA task by LLaMa-2-7B. The results are in Table~\ref{tab:tfqa-mc-lama2-7b}.
\begin{table}[h!]
\centering
\small
\small
\begin{tabular}{@{}p{0.2\linewidth}p{0.15\linewidth}p{0.15\linewidth}p{0.15\linewidth}@{}}
\toprule
\textbf{Method} & \textbf{MC1} & \textbf{MC2}  & \textbf{MC3} \\
\midrule
Baseline& 28.5 & 43.4 & 20.7 \\
 + Dola& 27.5 & 44.6  & 20.7  \\
+ Ours(0.5/24) & \bf29.0\up{0.6} & 46.9\up{3.5}& 22.1\up{1.4}  \\
+ Ours(0.5/26) & 28.3\down{0.2} & 45.3\up{1.9} & 21.2\up{0.5}  \\
+ Ours(single/26) & 27.1\downbad{1.4} & \bf61.1\up{17.7} & \bf32.9\up{12.2}  \\
\bottomrule
\end{tabular}
\caption{Multiple choices results of LLaMa-2-7B on TruthfulQA. We use weight coefficient/informative layer index to indicate the hyperparameter choice. For instance, 0.5/24 means we use $\alpha$=0.5 and use 24-th layer as the informative layer. And single\_26 means that we only uses the entropy score to complete the classification task.}
\label{tab:tfqa-mc-lama2-7b}
\end{table}

\section{Dataset Curation}
\label{append_sec:dataset_curation}

 We experiment with \counterfact~\citep{kevin2022locate} as a case study to showcase how inner representations tie with factuality. \counterfact~\cite{kevin2022locate} is a short-form QA dataset, each example $x$ is paired with a true answer $y_t$ and a constructed false answer $y_f$ (referred to as ``ground false'' in this paper). 
 Notably, all the examples in \counterfact~ contain annotations of knowledge triplets in each prompt, in the format of \textless subject, relation, object\textgreater. 
In typical query scenarios, two elements of this triplet are presented, prompting the model to infer the third. In \textsection\ref{sec:find1}, we will utilize these knowledge triplet annotations to study inner representations of specific locations. 

We aim to examine and compare the inner representations of the model in both cases of when the model produces factually correct and incorrect answers. 
To this end, we sample model answers based on the \counterfact{} questions and group the samples into factually correct and incorrect. 
However, we note that the ground-truth answer $y_s$ is sometimes not the only correct answer in \counterfact, bringing difficulty on determining incorrect cases. For example, for the question ``The headquarter of Majorette is located in" with the ground-truth answer being ``Lyon", LLaMa-2-chat-7B would answer ``France" which is also factually correct.
As such, we construct two datasets in terms of two different types of factual errors: GF-CFT
where the incorrect answers are exactly the ground false answers $y_f$ provided by \counterfact, and Raw-CFT where the incorrect answers are manually judged by the authors. 
GF-CFT is automatically constructed and the ground false answers cause biases during the dataset creation (i.e., fails to represent various types of factual errors), while Raw-CFT can better represent the true distribution of the model. 

Specifically, GF-CFT is constructed by firstly running inference of \texttt{LLaMA2-chat-7B} on CounterFact using 2-shot prompt. Then we obtain all the cases where the generated text is exactly the ground truth, where there are 325 samples. 
Next we randomly sample 700 cases where the generated text is exactly the ground false. 
Raw-CFT is constructed by firstly randomly sampled 1000 cases in CounterFact and inference by \texttt{LLAMA2-7B-chat}. Then the authors annotate them and keep 364 of them that is factually correct or incorrect (the remaining 636 samples generate irrelevant content). 

\section{Hyperparameter Generalization}

\begin{table*}[ht]
\centering
\small
\resizebox{1.0 \linewidth}{!}{
\begin{tabular}{llllllll}
\toprule
\multirow{2}{*}{\textbf{Model}} & \multicolumn{2}{c}{\textbf{TriviaQA}}  & \multicolumn{2}{c}{\textbf{HotPotQA}}  & \multicolumn{2}{c}{\textbf{NQ}}\\
\cmidrule(lr){2-8}
  &  \textbf{Exact Match} & \textbf{F1 score} &  \textbf{Exact Match}  & \textbf{F1 score}&  \textbf{Exact Match}  & \textbf{F1 score}\\
\midrule
LLaMa2-7B-chat  & 44.4&44.3 & 19.6&20.1&21.8&20.4 \\
 + ITI~\citep{li2023inference} & \bf 46.5&\bf46.5 & 19.7&19.7&\bf23.5&\bf21.5 \\
 + Dola &  45.2&45.3 & \bf 20.4&\bf 21.3 &22.8&21.2 \\
 + Ours & 45.0\up{0.6}&44.4\up{0.1} & 20.2\up{0.6}&20.8\up{0.7}&22.1\up{0.3}&21.0\up{0.6}\\
\midrule
LLaMa2-13B-chat  & 63.0&60.9 &23.8&21.7 &33.1&28.9 \\
  + ITI~\citep{li2023inference}  & 63.0 & 60.9 & 23.8&21.7&33.1&28.9 \\
  + Dola  & 63.2&61.5&24.5&23.2  &34.6&31.2 \\
 + Ours  & \bf64.4\up{1.4}&\bf62.7\up{1.8}&\bf24.9\up{1.4}&\bf23.3\up{0.7} &\bf35.8\up{2.7}&\bf32.4\up{3.5} \\
\midrule
LLaMa2-70B-chat  &73.3&68.4  &30.2&25.5 &40.7&34.1 \\
+ ITI~\citep{li2023inference} & 73.4 & 68.5 & 30.2& 25.6&40.7&34.1 \\
 + Dola &  74.1&72.3 & \bf31.2&\bf29.0&41.9&36.2 \\
 + Ours & \bf74.4\up{1.1}&\bf73.2\up{4.8} &30.7\up{1.3}&27.4\up{1.1} &\bf42.3\up{1.6}&\bf37.4\up{3.3} \\
\bottomrule
\end{tabular}}
\vspace{-5pt}
\caption{Open-ended generation results on TriviaQA, HotPotQA and Natural Questions (metrics are in $\times 10^{-2}$). Different from Table~\ref{tab:combined}, the hyperparameters of all baselines and our approach here are selected based on TruthfulQA dataset rather than on the respective benchmark dataset, representing an out-of-domain evaluation setting. The best-performing methods are in bold. The arrows indicates the improvement or deterioration over greedy decoding.}
\label{tab:qa_ood}
\vspace{-5pt}
\end{table*}

\paragraph{Parameter setting} Our method involves two key hyperparameters: the index of the informative layer and the weight coefficient. 
To test the generalization ability of our method and ensure uniformity in our experimental outcomes, we standardized the parameters for models of equivalent size across all benchmarks. The two hyperparameters are optimized on the TruthfulQA Multiple Choice task.

For the LLaMa2-7B-chat model, we set the informative layer as 26 and the alpha as 0.5. For the LLaMa2-13B-chat model, we set the informative layer as 34 and the alpha as 0.8. For the LLaMa2-70B-chat model, we set the informative layer as 70 and the alpha as 1.

\section{Inference Efficiency}
\begin{figure}[ht]
  \centering
  \includegraphics[width=0.42\textwidth]{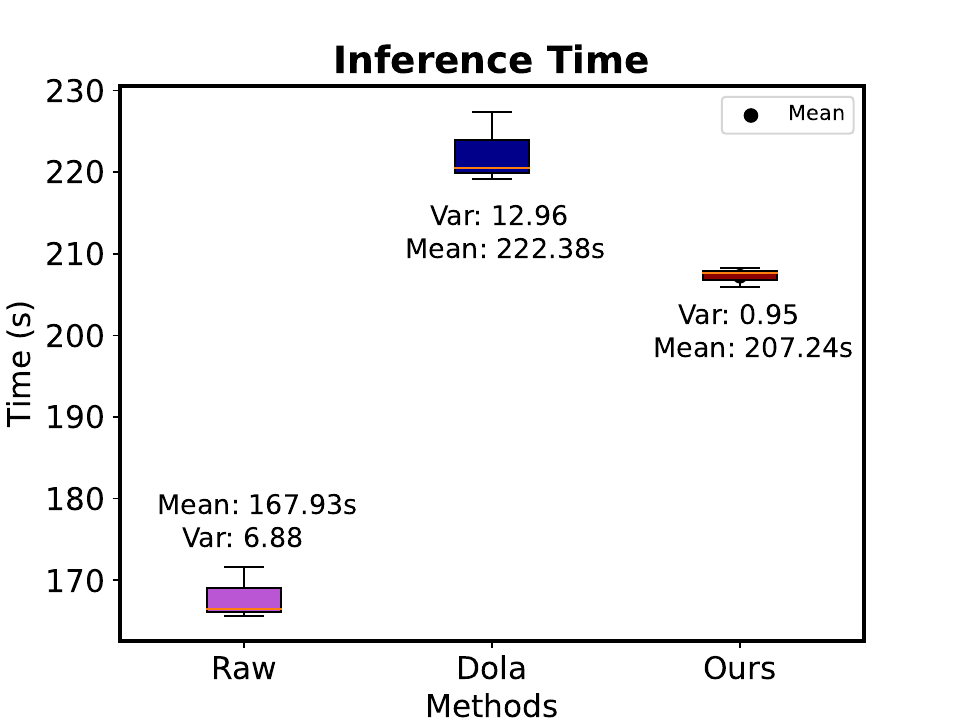}
 \caption{Comparison of Inference time on 722 samples from Natural Questions (we randomly sample 20\% of the validation set) using LLaMA-2-chat-7B model on a single NVIDIA Tesla A800 80GB GPU.}
  \label{fig:inference_time}
  \vspace{-5mm}
\end{figure}

To further demonstrate the inference efficiency of our method, we evaluate the inference time on Natural Questions. We compare the inference time with Dola.
\figureautorefname{~\ref{fig:inference_time}} shows that our method reduces inference time by 7.3\% compared to Dola while increases the inference time by 23.4\% compared to the greedy decoding. Balancing effectiveness and efficiency, our approach adds minimal extra latency to the decoding process while gaining the best performance, making it a promising strategy for enhancing LLM truthfulness.

\begin{algorithm}
    \caption{Activation Decoding for Text Generation}
    \label{alg:activate_decoding}
\begin{algorithmic}[1]
    \State {\bfseries Input:} Prompt prefix $\mathcal{C}=\{v_1\dots v_h\}$, language model $\mathcal{M}$ with vocabulary $\mathcal{V}$, informative layer $l$ and hyperparameter $\alpha$, max token length $T$, threshold $\tau$.
    \State {\bfseries Output:} Continuation $\mathcal{G}=\{x_{h+1}\dots x_{h+n}\}$
    \State $\mathcal{G}\gets\{\}$
    \State \linecomment{Use LLM to transform in-context tokens, saving hidden states at layer $l$}
    \State Use LLM $\mathcal{M}$ to transform the in-context tokens and save the sequence of hidden states $\{\mathbf{x}_1^{l}, \dots, \mathbf{x}_{h}^{l}\}$
    \State \linecomment{Pre-compute entropy for all tokens in $\mathcal{V}$}
    \For{$v_t \in \mathcal{V}$}
        \For{$v_j \in \mathcal{C}$}
            \State $P(v_{t} \mid v_{\leq j}) = \mathrm{softmax}\bigl(\phi(\mathbf{x}_{j}^l)\bigr)_{v_{t}}$ \inlinecomment{Compute activation score}
        \EndFor
        \State $E(v_t |v_{\leq h}) = -\sum_{i=1}^{h} P(v_i \mid v_{\leq i}) \log P(v_i \mid v_{\leq i})$ \inlinecomment{Compute entropy}
    \EndFor
    \State \inlinecomment{Generate tokens using activation decoding}
    \State $t = h+1$
    \While{stop token not generated \textbf{and} $t \leq T+h$}
        \State $q_v = \mathrm{softmax}\bigl(\phi(\mathbf{x}_{t}^l)\bigr)$ \inlinecomment{Next token probability distribution}
        \For{$v_t \in \{ v_i | q_v(v_i) \geq \tau \, \underset{w}{\max} q_v(w) \}$}
            \State $P_q(v_t \mid v_{<t}) = e^{ - \alpha E(v_t \mid v_{\leq h}) } P_q(v_t \mid  v_{<t})$ \inlinecomment{Adjust probability}
        \EndFor
        \State $x_t=\textbf{argmax}_{v\in \mathcal{V}}P_q(v|v_{<t})$
        \State $\mathcal{G}\gets\mathcal{G}\cup\{x_t\}$
    \EndWhile
\end{algorithmic}
\end{algorithm}

\end{document}